\documentclass[runningheads]{llncs}
\usepackage{graphicx}
\usepackage{amsmath}
\usepackage{amssymb}
\usepackage{booktabs}
\usepackage{pifont}
\usepackage{tabularx}
\usepackage[table]{xcolor}
\usepackage{amssymb}

\begin{document}

\title{A Spatial-Frequency Aware Multi-Scale Fusion Network for Real-Time Deepfake Detection\thanks{Supported by the Taishan Scholars Program (NO.tspd20240814) and by Key Laboratory of Computing Power Network and Information Security, Ministry of Education under Grant No.2023ZD033.}}
\titlerunning{Real-Time Deepfake Detection with Spatial-Frequency Fusion}
%
\author{Libo Lv\inst{1} \and
Tianyi Wang\inst{2}\thanks{Corresponding author.} \and
Mengxiao Huang\inst{1} \and
Ruixia Liu\inst{1} \and
Yinglong Wang\inst{3}}
\authorrunning{Lv et al.}
%
\institute{
Qilu University of Technology (Shandong Academy of Sciences); Shandong Artificial Intelligence Institute, China \\
\email{libo75774@gmail.com, huangmx001104@163.com, liurx@sdas.org} 
\and
School of Computing, National University of Singapore, Singapore \\
\email{wangty@nus.edu.sg}
\and
Key Laboratory of Computing Power Network and Information Security, Ministry of Education; Qilu University of Technology (Shandong Academy of Sciences), China \\
\email{wangyinglong@qlu.edu.cn}}

\maketitle
%
\begin{abstract}
With the rapid advancement of real-time deepfake generation techniques, forged content is becoming increasingly realistic and widespread across applications like video conferencing and social media. Although state-of-the-art detectors achieve high accuracy on standard benchmarks, their heavy computational cost hinders real-time deployment in practical applications. To address this, we propose the Spatial-Frequency Aware Multi-Scale Fusion Network (SFMFNet), a lightweight yet effective architecture for real-time deepfake detection. We design a spatial-frequency hybrid aware module that jointly leverages spatial textures and frequency artifacts through a gated mechanism, enhancing sensitivity to subtle manipulations. A token-selective cross attention mechanism enables efficient multi-level feature interaction, while a residual-enhanced blur pooling structure helps retain key semantic cues during downsampling. Experiments on several benchmark datasets show that SFMFNet achieves a favorable balance between accuracy and efficiency, with strong generalization and practical value for real-time applications.

\end{abstract}
\section{Introduction}
Recent advances in AI and deep learning have greatly improved deepfake technology, enabling highly realistic video and audio synthesis ~\cite{Simswap,StarGANv2,APSwap}. While benefiting for film production and virtual entertainment, these developments raise privacy, ethical, and security concerns~\cite{DeepfakeSurvey2024Wang}. With better hardware and optimized algorithms, Real-Time Deepfakes~\cite{RTDFs} (RTDFs) now allow instant manipulation of live video and audio for broadcasts, video calls, and virtual reality. Researchers are working to overcome challenges like latency, synchronization, accuracy, and security through network optimization, hardware acceleration, and adaptive learning. As RTDFs expand applications in entertainment and remote communication, addressing risks and ensuring data security and ethical use remain crucial.

In recent years, deepfake detection has made significant progress, largely driven by the introduction of convolutional neural network-based approaches. Numerous studies have demonstrated that deep models can effectively identify subtle artifacts and manipulation traces introduced during forgery, achieving high detection accuracy and reasonable generalization across benchmark datasets such as FaceForensics++~\cite{FF++} (FF++) and Celeb-DF~\cite{Celeb-DF_v2}. However, most of these models rely on deep and complex architectures with high computational overhead, making them difficult to deploy in resource-constrained or latency-sensitive environments, and thereby limiting their applicability in real-time deepfake detection scenarios.

To support real-time applications, recent works have explored lightweight detection models by reducing input resolution, pruning backbones, or designing efficient architectures~\cite{BDD,LaDeDa}. While these approaches improve inference speed, they often suffer from limited representational capacity, reduced sensitivity to subtle artifacts, or loss of spatial and frequency details crucial for robust detection. Furthermore, many existing methods treat spatial and frequency inconsistencies independently, neglecting their joint contribution in revealing forgery clues. The key challenge, therefore, lies in designing real-time detection frameworks that maintain high accuracy and generalization while operating under constrained computational budgets. Addressing these issues, we propose a spatial-frequency hybrid aware network that explicitly models the interaction between spatial textures and frequency artifacts, enabling more accurate and efficient real-time deepfake detection.

Our method strikes an effective balance between performance and computational cost, outperforming most existing state-of-the-art approaches. We conducted extensive experiments on six widely used public benchmark datasets to comprehensively validate the effectiveness and efficiency of the proposed method. The main contributions of our work are summarized as follows:

\vspace{0.5em}

\noindent
\begin{tabularx}{\linewidth}{@{}cX@{}}
\textbullet & We propose a spatial–frequency hybrid aware module that enhances forgery region perception for real-time applications by fusing wavelet features and coordinate attention into a dynamic gating map. \\
\textbullet & The token-selective cross attention module is introduced, enabling efficient cross-scale feature interaction through token sampling and fusion, improving forgery feature alignment and discrimination. \\
\textbullet & A residual downsampling module based on blur pooling is designed to better preserve structural and edge details while mitigating aliasing during multi-scale feature integration. \\
\end{tabularx}

\section{Related Work}
\subsection{Conventional Deepfake Detection}

Deepfake detection has evolved significantly over time. Early approaches relied on low-level physical cues, such as unnatural facial movements \cite{inconsistencies} and compression artifacts \cite{JPEG compression artifacts}. More recent methods employ deep neural networks, particularly convolutional neural networks (CNNs) and transformer architectures, to capture complex spatial and temporal patterns. For instance, recent methods have introduced multi-head relative interaction mechanisms to analyze subtle noise patterns \cite{NoiseDF2023Wang}. In addition, the combination of CNN-based feature extraction and the global information modeling capabilities of transformers \cite{DCPT2023Wang} has significantly improved detection accuracy. With the emergence of diffusion models and large-scale generative models~\cite{ASAP}, the visual fidelity of forged content has surpassed traditional GAN-based methods. However, despite the remarkable performance of current detection models, their high computational complexity poses a challenge for real-time deployment, particularly in applications such as live content moderation and video conferencing.

\subsection{Lightweight Detection Models }
\begin{sloppypar}

In recent years, researchers have focused on model compression, lightweight architectures, and attention mechanisms to improve detection efficiency without sacrificing accuracy. Early efforts centered on pruning, which simplifies conventional networks to reduce computational complexity. Efficient backbones, such as AlexNet~\cite{AlexNet}, provided strong feature extraction while maintaining lower complexity. Attention mechanisms, exemplified by GoogLeNet~\cite{GoogLeNet}, further enhanced the network's ability to capture salient information with minimal parameter overhead, boosting performance in challenging detection scenarios. In deepfake detection, attention-based approaches have proven effective for extracting fine-grained features~\cite{extract fine-grained features} and localizing manipulated regions~\cite{locate manipulated regions}. Spatial and channel attention modules are particularly adept at emphasizing anomalies, while cross-layer attention facilitates multi-scale feature fusion. However, existing lightweight attention modules, such as CBAM~\cite{CBAM}, still incur significant computational costs, limiting their use in ultra-lightweight frameworks. Moreover, traditional fusion strategies often struggle to model complex feature interactions, further hindering detection performance.
\end{sloppypar}

\section{Proposed method}
\begin{figure}
\centering
\includegraphics[width=\textwidth]{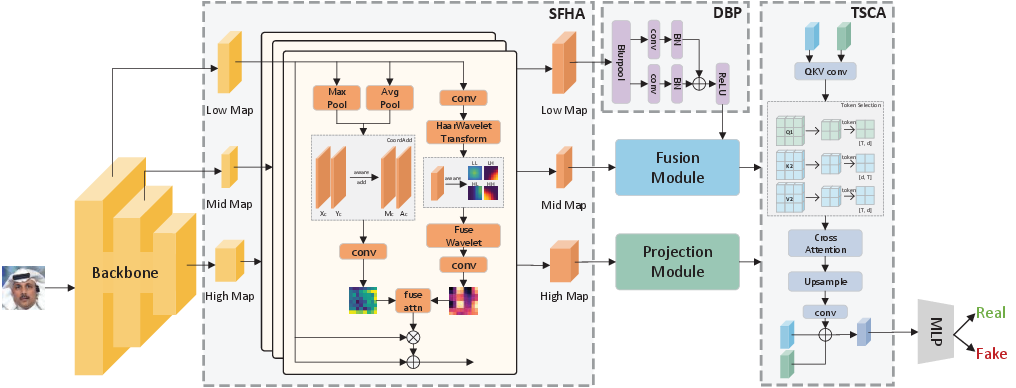}
\caption{Overall architecture of the proposed lightweight forgery detection model. CNN backbone extracts multi-level features, enhanced by spatial-frequency hybrid aware module to highlight forgery regions. Low-level features are downsampled and fused with mid-level features, then refined through token-selective cross-attention with high-level features. Finally, global pooling and a multilayer perceptron classify forgeries, balancing accuracy and efficiency.} \label{fig1}
\end{figure}

The architecture of our proposed method is shown in Fig.~\ref{fig1}. We employ a CNN backbone to extract multi-scale features and enhance artifacts using a spatial-frequency hybrid aware module. Low-level features are downsampled and fused with mid-level features, while high-level features are mapped into a compressed space. Selective token cross-attention efficiently models artifacts by linking fused mid- and low-level features with mapped high-level features. Finally, global pooling and an adaptive MLP classifier produce a binary classification result.

\subsection{Spatial-Frequency Hybrid Aware Module}
To highlight forged regions, we propose the Spatial-Frequency Hybrid Aware (SFHA) Module, which extracts wavelet-domain~\cite{wavelet} features and spatial attention maps in parallel. These are fused into a dynamic gating map that emphasizes critical areas of the input features. Given a low-level feature map $\mathrm{X}_{\text{low}} \in \mathbb{R}^{C \times H \times W}$ from a CNN backbone, the wavelet branch first computes the frequency-aware attention map $\mathrm{M}_{\mathrm{freq}}$ as follows:
\begin{equation}
    \mathrm{M}_{\mathrm{freq}}=\mathcal{K}_f\left(\mathcal{U}\left(\mathrm{ReLU}\left(\mathcal{\tilde{K}}_f\left(\mathrm{Haar}(\mathrm{X}_{\text{low}})\right)\right)\right)\right),
\end{equation}
where $\mathcal{K}_f$ denotes 2D convolution in the frequency domain, $\mathcal{U}$ denotes upsampling, $\tilde{\mathcal{K}}_f$ denotes 3D convolution, and $\mathrm{Haar}(\cdot)$ denotes channel-wise Haar wavelet transform, producing four sub-bands. The spatial branch then yields the spatial attention map $\mathrm{M}_{\mathrm{spa}}$ as follows:
\begin{equation}
   \mathrm{M}_{\text{spa}} = \mathcal{K}_s \cdot (\mathrm{CoordAdd}([\frac{1}{C}\sum^C_{c=1}\mathrm{X}_{\text{low}},\mathrm{max^C_{c=1}}\mathrm{X}_{\text{low}}])),
\end{equation}
where $\mathcal{K}_s$ denotes 2D convolution in the spatial domain, and $\mathrm{CoordAdd}(\cdot)$ denotes the concatenation of normalized coordinate channels as extra positional cues. The notation $[,]$ represents channel-wise concatenation of two feature maps. The two attention maps are then adaptively fused to produce the low-level spatial–frequency gating $\mathrm{G}_{\text{low}}$ as follows:
\begin{equation}
    \mathrm{G}_{\text{low}} = \sigma(\mathcal{K}_g\cdot[\alpha\cdot \mathrm{M}_{\mathrm{freq}},\beta\cdot\mathrm{M}_{\text{spa}}]),
\end{equation}
where $\mathcal{K}_g$ denotes 2D convolution, $\sigma$ denotes the sigmoid activation, and $\alpha,\beta$ are learnable weights for fusing frequency- and spatial-domain attention. The resulting hybrid gating $\mathrm{G}_{\text{low}}$ is then applied to the input feature $\mathrm{X}_{\text{low}}$, and the enhanced output $\mathrm{Y}_{\text{low}}$ is computed with a residual connection as follows:
\begin{equation}
    \mathrm{Y}_{\text{low}} = \mathrm{X}_{\text{low}} +\mathrm{G}_{\text{low}} \odot \mathrm{X}_{\text{low}},
\end{equation}
where $\odot$ denotes element-wise multiplication. To account for varying receptive fields and semantics at different feature levels, we use different convolution kernel sizes in the spatial branch. This enables multi-scale spatial feature extraction and yields the spatial–frequency enhanced features $\mathrm{Y}_{\text{low}}$, $\mathrm{Y}_{\text{mid}}$, and $\mathrm{Y}_{\text{high}}$.

\subsection{Downsample with BlurPool Module}
To reduce aliasing and better retain structure and edge details during downsampling, we introduce the Downsample with BlurPool (DBP) module. The input feature $\mathrm{Y}_{\text{low}}$ is first smoothed by blur pooling~\cite{BlurPool}, then passed through convolution and ReLU to obtain the main semantic representation $\mathrm{D}_{\text{mian}}$ as follows:
\begin{equation}
    \mathrm{D}_{\text{main}} = \mathrm{ReLU}(\mathrm{BN}(\mathcal{K}_m(\mathcal{B}(\mathrm{Y}_{\text{low}})))),
\end{equation}
where $\mathcal{K}_m$ denotes convolution, and $\mathcal{B}$ denotes blur pooling. To reduce information loss, we introduce a residual path that applies the same blur and linear projection to the original input $\mathrm{Y}_{\text{low}}$ as follows:
\begin{equation}
    \mathrm{D}_{\mathrm{res}}=\mathrm{BN}(\mathcal{K}_r(\mathcal{B}(\mathrm{Y}_{\text{low}}))),
\end{equation}
this path acts as an information compensation branch. Unlike the previous design, we apply a standard $1 \times 1$ convolution $\mathcal{K}_r$ to the blurred features to capture texture details, improving training stability and retaining low-level information. Finally, we sum the main and residual paths and apply ReLU to produce the downsampled output as follows:
\begin{equation}
    \mathrm{D}_{\text{low}}=\mathrm{ReLU}(\mathrm{D}_{\mathrm{main}}+\mathrm{D}_{\mathrm{res}}).
\end{equation}

For improved multi-level fusion, we apply the Projection Module to high-level features for mapping, and use the Fusion Module to integrate low- and mid-level features, as detailed below:
\begin{equation}
    \mathrm{F}_{\mathrm{proj}}= \mathcal{P}(\mathrm{Y}_{\text{high}}), 
\end{equation}
\begin{equation}
    \mathrm{F}_{\mathrm{fuse}}=\mathcal{F}(\mathrm{D}_{\text{low}},\mathrm{Y}_{\text{mid}}),
\end{equation}
where $\mathcal{P}$ denotes the Projection Module, comprising convolutional, normalization, and activation layers. $\mathcal{F}$ denotes the Fusion Module, which concatenates feature maps followed by convolutional layers.

\subsection{Token-Selective Cross Attention Module}
We propose the Token-Selective Cross Attention (TSCA) module to enhance low-level details with high-level semantics while reducing computation. TSCA compresses spatial information, selects key tokens, and performs cross-attention to compute Query, Key, and Value representations as follows:
\begin{equation}
\left\{
\begin{aligned}
\mathrm{Q}_1 &= \mathcal{K}_q(\mathrm{F}_{\mathrm{fuse}}),\\
\mathrm{K}_2 &= \mathcal{K}_k(\mathrm{F}_{\mathrm{proj}}),\\
\mathrm{V}_2 &= \mathcal{K}_v(\mathrm{F}_{\mathrm{proj}}),
\end{aligned}
\right.
\end{equation}
where $\mathrm{Q}_1, \mathrm{K}_2, \mathrm{V}_2 \in \mathbb{R}^{d \times H \times W}$. $\mathcal{K}_q$, $\mathcal{K}_k$, and $\mathcal{K}_v$ denote three independent convolutional layers that generate the query, key, and value feature maps, respectively. The parameter $d$ is the compressed channel dimension. To further reduce computation, the Token Selection mechanism applies spatial average pooling to $\mathrm{Q}_1$, $\mathrm{K}_2$, and $\mathrm{V}_2$, producing token representations $\tilde{\mathrm{Q}}_1$, $\tilde{\mathrm{K}}_2$, and $\tilde{\mathrm{V}}_2$ as follows:
\begin{equation}
    \tilde{\mathrm{Q}}_1,\tilde{\mathrm{K}}_2,\tilde{\mathrm{V}}_2 = \mathcal{R}(\frac{1}{C}\sum^C_{c=1}(\mathrm{Q}_1,\mathrm{K}_2,\mathrm{V}_2)),
\end{equation}
where $\tilde{\mathrm{K}}_2 \in \mathbb{R}^{d \times T}$ and $\tilde{\mathrm{Q}}_1, \tilde{\mathrm{V}}_2 \in \mathbb{R}^{T \times d}$, $T = t \times t$ is the number of selected tokens, with $t$ the pooling size. The operator $\mathcal{R}$ denotes reshaping. The standard scaled dot-product attention is then computed as follows:
\begin{equation}
    \mathrm{Z} = (\mathrm{Softmax}(\frac{\tilde{\mathrm{Q}}_1 \cdot \tilde{\mathrm{K}}_2^{T}}{\sqrt{d}})  \cdot \tilde{\mathrm{V}}_2,
\end{equation}
where $\mathrm{Z}$ denotes the semantically enriched token representation, $d$ is the dimensionality of the query and key embeddings. It is reshaped to spatial size ($t \times t$), upsampled to the original feature resolution, and passed through a $1 \times 1$ convolution to adjust channels. Finally, it is fused with the original low-level features to produce the output $\mathrm{F}_{\text{out}}$ as follows:
\begin{equation}
    \mathrm{F}_{\text{out}} = \alpha \cdot \mathcal{K}(\mathrm{Upsample}(\mathcal{R}(\mathrm{Z}))) + (1-\alpha) \cdot \frac{\mathrm{F}_{\mathrm{fuse}} + \mathrm{F}_{\mathrm{proj}}}{2},
\end{equation}
where $\alpha$ is a learnable scalar weight used for residual balancing, and $\mathcal{K}$ represents the convolution operation.
\subsection{Loss functions}
We train the model with a weighted combination of cross-entropy loss, gated regularization loss, and consistency loss. Cross-entropy guides real or fake classification, the regularization term controls attention activation, and consistency loss stabilizes predictions under variations.

\subsubsection{Classification Loss.} We use standard binary cross-entropy loss $\mathcal{L}_{\mathrm{CE}}$ to supervise the classification of real and fake samples. This loss encourages the model to produce high confidence for correct predictions.

\subsubsection{Gated Regularization Loss.}
To constrain the response patterns between attention maps at different scales and image labels, we propose the gated regularization loss as follows:
\begin{equation}
    \mathcal{L}_{\mathrm{gate}}=\frac{1}{3}\left(\ell(\mathrm{G}_{\mathrm{low}},y)+\ell(\mathrm{G}_{\mathrm{mid}},y)+\ell(\mathrm{G}_{\mathrm{high}},y)\right),
\end{equation}
where $\mathrm{G}_{\mathrm{low}}, \mathrm{G}_{\mathrm{mid}}, \mathrm{G}_{\mathrm{high}}$ represent the globally averaged pooled vectors of the low-, mid-, and high-level spatial-frequency hybrid gating, respectively. The term $y \in \{0,1\}^{N \times 1}$ denotes the labels for positive and negative samples, and $\ell (\cdot, \cdot)$ represents the mean squared error loss as follows:
\begin{equation}
    \ell(a,b)=\frac{1}{N}\sum_{i=1}^N(a_i-b_i)^2,
\end{equation}
this loss encourages the gated attention maps of positive samples to exhibit higher response values, while the responses for negative samples tend to be lower.
\subsubsection{Consistency Loss.}
We further introduce a consistency loss to maintain the stability and coherence of attention responses across scales. It is defined as follows:
\begin{equation}
    \mathcal{L}_{\mathrm{cons}}=\frac{1}{3}\left(\ell(\mathrm{G}_{\mathrm{low}},\mathrm{G}_{\mathrm{mid}})+\ell(\mathrm{G}_{\mathrm{low}},\mathrm{G}_{\mathrm{high}})+\ell(\mathrm{G}_{\mathrm{mid}},\mathrm{G}_{\mathrm{high}})\right),
\end{equation}
this loss ensures that the model consistently identifies discriminative regions at different scales. This contributes to improved overall robustness.
\subsubsection{Total Loss.}
The final total loss function is defined as the weighted sum of the three components:
\begin{equation}
    \mathcal{L}_{\mathrm{total}}=\lambda_{\mathrm{CE}}\cdot\mathcal{L}_{\mathrm{CE}}+\lambda_{\mathrm{gate}}\cdot\mathcal{L}_{\mathrm{gate}}+\lambda_{\mathrm{cons}}\cdot\mathcal{L}_{\mathrm{cons}},
\end{equation}
where $\lambda_{\mathrm{CE}}, \lambda_{\mathrm{gate}}, \lambda_{\mathrm{cons}}$ are the weights that balance the classification loss, gated regularization loss, and consistency loss, respectively.
\section{Experients}
\subsection{Experimental Settings}
\subsubsection{Datasets.}
The FF++~\cite{FF++} dataset is used for deepfake detection, containing real and manipulated face videos produced with four different forgery techniques, making it one of the most widely adopted benchmarks in this field. For evaluating the generalization performance of deepfake detection models, we also consider several well-established datasets: the DeepFake Detection~\cite{DFD} (DFD), the Celeb-DF v2~\cite{Celeb-DF_v2} (CDF2), the DeepFake Detection Challenge Preview~\cite{DFDCP} (DFDCP), the DeepFake Detection Challenge~\cite{DFDC} (DFDC), and the UAD Fake Video~\cite{UADFV} (UADFV). To maintain consistency with previous work, we strictly follow the image cropping and data preprocessing protocols defined by DeepfakeBench~\cite{DeepfakeBench}.
\subsubsection{Implementation Details.}
The experiments are conducted using the PyTorch Lightning framework for model training, validation, and testing, running entirely on NVIDIA V100 GPUs. To ensure stability and reproducibility, all configurations are kept fixed throughout the training process. The input image size is set to 256×256, with a batch size of 32. The Adam optimizer is used with a learning rate of 0.0002. The model is optimized on the training dataset with the loss weights set as $\lambda_{\mathrm{CE}}=1.0$, $\lambda_{\mathrm{gate}}=0.3$ and $\lambda_{\mathrm{cons}}=0.2$. Its performance is evaluated on the validation set after each epoch using the Area Under the Curve (AUC). Computational complexity and model size are assessed by reporting the number of Floating Point Operations (FLOPs) and parameters. After training, the model is tested on a separate test set to evaluate its generalization ability. The checkpoint achieving the highest AUC on the validation set is saved and used for final evaluation on the test set, ensuring the model’s robustness and practical applicability.

\subsection{Evaluations}

\begin{table}[ht]
\centering
\caption{Compares the performance of various detection methods within the same dataset.The best results are highlighted in bold.}
\renewcommand{\arraystretch}{1.2}
\setlength{\tabcolsep}{5pt}
\begin{tabular}{lccccc|c}
\hline
\textbf{Model} & \textbf{FF++} & \textbf{FF-DF} & \textbf{FF-F2F} & \textbf{FF-FS} & \textbf{FF-NT} & \textbf{Avg.} \\
\hline
RegNet\cite{RegNet} & 0.9806 & 0.9903 & 0.9845 & 0.9892 & 0.9715 & 0.9833 \\
GoogLeNet\cite{GoogLeNet} & 0.9810 & 0.9927 & 0.9877 & 0.9884 & 0.9554 & 0.9810 \\
AlexNet\cite{AlexNet} & 0.9415 & 0.9632 & 0.9623 & 0.9576 & 0.8828 & 0.9414 \\
ResNet18\cite{Resnet18} & 0.9810 & 0.9904 & 0.9860 & 0.9884 & 0.9592 & 0.9810 \\
CNN-Aug\cite{CNN-Aug} & 0.8493 & 0.9048 & 0.8788 & 0.9026 & 0.7313 & 0.8533 \\
Xception\cite{Xception} & 0.9637 & 0.9799 & 0.9785 & 0.9833 & 0.9385 & 0.9687 \\
Capsule\cite{Capsule} & 0.8421 & 0.8669 & 0.8634 & 0.8734 & 0.7804 & 0.8452 \\
FWA\cite{FWA} & 0.8765 & 0.9210 & 0.9000 & 0.8843 & 0.8120 & 0.8787 \\
X-ray\cite{X-rey} & 0.9592 & 0.9794 & 0.9872 & 0.9871 & 0.9290 & 0.9683 \\
FFD\cite{FFD} & 0.9624 & 0.9803 & 0.9784 & 0.9853 & 0.9306 & 0.9674 \\
UCF\cite{UCF} & 0.9705 & 0.9883 & 0.9840 & \textbf{0.9896} & 0.9441 & 0.9753 \\
F3Net\cite{F3Net} & 0.9635 & 0.9793 & 0.9796 & 0.9844 & 0.9354 & 0.9684 \\
SPSL\cite{SPSL} & 0.9610 & 0.9781 & 0.9754 & 0.9829 & 0.9299 & 0.9654 \\
\hline
\textbf{ours} & \textbf{0.9868} & \textbf{0.9949} & \textbf{0.9884} & 0.9890 & \textbf{0.9748} & \textbf{0.9867} \\
\hline
\end{tabular}
\label{tab:1}
\end{table}

\begin{figure}
\centering
\includegraphics[width=\textwidth]{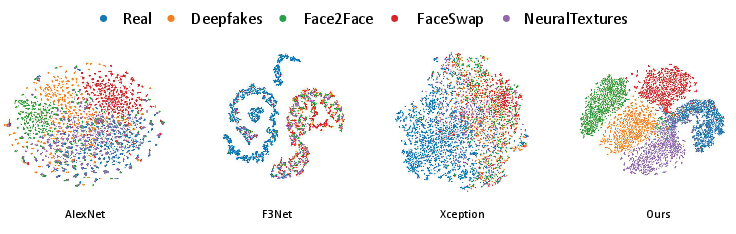}
\caption{t-SNE visualization for each detector. These detectors are trained and tested on FF++.} \label{fig2}
\end{figure}

Given the real-time requirements of our task, we focus on detectors with low inference costs. Large-scale vision-language and diffusion-based detectors, such as ASAP~\cite{ASAP}, often exceed 100 GFLOPs and billions of parameters, making them impractical for latency-sensitive applications. In contrast, spatial- and frequency-based detectors like UCF~\cite{UCF} and X-ray~\cite{X-rey} operate under 10 GFLOPs with fewer than 100 million parameters, offering competitive accuracy and high efficiency. We thus select recent detectors with similarly low computational footprints as baselines for a fair comparison in real-time scenarios.

\begin{table}[htbp]
\centering
\caption{Examines the model’s generalization ability across different datasets.}
\renewcommand{\arraystretch}{1.2}
\resizebox{\textwidth}{!}{
\begin{tabular}{lcccccc|ccc}
\toprule
\textbf{Model} & \textbf{FF++} & \textbf{DFD} & \textbf{CDF2} & \textbf{DFDCP} & \textbf{DFDC} & \textbf{UADFV} & \textbf{Avg.} & \textbf{FLOPs(G)} & \textbf{Params(M)} \\
\midrule
RegNet\cite{RegNet}     & 0.9806 & 0.8387 & 0.7173 & 0.7105 & 0.6902 & 0.7965 & 0.7889 & 2.11 & 9.19 \\
GoogLeNet\cite{GoogLeNet}  & 0.8321 & 0.7321 & 0.6967 & 0.7259 & 0.6505 & 0.8785 & 0.7526 & 1.96 & 13.01 \\
AlexNet\cite{AlexNet}    & 0.9415 & 0.7787 & 0.7538 & 0.7127 & 0.6918 & 0.8637 & 0.7903 & 1.32 & 61.12 \\
ResNet18\cite{Resnet18}   & 0.9810 & 0.8030 & 0.7731 & 0.7039 & 0.6736 & 0.8653 & 0.7999 & 2.37 & 11.69 \\
CNN-Aug\cite{CNN-Aug}    & 0.8493 & 0.6464 & 0.7027 & 0.6170 & 0.6361 & 0.8739 & 0.7209 & 1.93 & 11.73 \\
Xception\cite{Xception}   & 0.9637 & 0.8163 & 0.7365 & 0.7374 & 0.7077 & 0.9379 & 0.8165 & 5.98 & 20.81 \\
Capsule\cite{Capsule}   & 0.8421 & 0.6841 & 0.7472 & 0.6568 & 0.6465 & 0.9078 & 0.7474 & 19.80 & 20.21 \\
FWA\cite{FWA}       & 0.8765 & 0.7403 & 0.6680 & 0.6373 & 0.6132 & 0.8539 & 0.7315 & 5.98 & 20.94 \\
X-ray\cite{X-rey}      & 0.9592 & 0.7655 & 0.6786 & 0.6942 & 0.6326 & 0.8989 & 0.7715 & 8.93 & 96.50 \\
FFD\cite{FFD}        & 0.9624 & 0.8024 & 0.7435 & 0.7426 & 0.7029 & 0.9450 & 0.8164 & 6.01 & 22.37 \\
UCF\cite{UCF}        & 0.9705 & 0.8074 & 0.7527 & 0.7594 & 0.7179 & \textbf{0.9528} & 0.8267 & 10.53 & 47.64 \\
F3Net\cite{F3Net}      & 0.9635 & 0.7975 & 0.7352 & 0.7354 & 0.7021 & 0.9347 & 0.8114 & 6.19 & 21.32 \\
SPSL\cite{SPSL}       & 0.9610 & 0.8122 & 0.7650 & 0.7408 & 0.7040 & 0.9424 & 0.8209 & 7.65 & 20.82 \\
\hline
\textbf{ours} & \textbf{0.9868} & \textbf{0.8781} & \textbf{0.8111} & \textbf{0.8603} & \textbf{0.7307} & 0.9420 & \textbf{0.8682} & \textbf{1.27} & \textbf{6.64} \\
\bottomrule
\end{tabular}
}
\label{tab:2}
\end{table}

To comprehensively evaluate the forgery detection performance of various models within a unified training domain, we conducted in-domain testing on the FaceForensics++ dataset, with results presented in Table~\ref{tab:1}. Our observations indicate that the proposed method achieves superior performance across most forgery subsets, attaining an average AUC of 0.9867, surpassing multiple state-of-the-art approaches. Notably, it achieves an AUC of 0.9748 on the complex FF-NT subset, demonstrating strong adaptability and robustness.

We used t-SNE to visualize the feature distributions learned by different models, as shown in Fig.~\ref{fig2}. AlexNet and Xception show significant overlap between real and fake samples, indicating limited discriminative ability. F3Net performs better but still mixes different forgery types. In contrast, our method produces well-separated clusters, clearly distinguishing real and various fake sources, demonstrating superior feature representation.

Table~\ref{tab:2} compares our method with state-of-the-art models on six datasets. Compared to UCF, our method achieves AUC improvements of 1.63\% on FF++, 7.07\% on DFD, 6.76\% on CDF2, 10.09\% on DFDCP, and 1.28\% on DFDC. On UADFV, our AUC is 0.9420, slightly lower than UCF's 0.9528. Overall, our average AUC reaches 0.8682, which is 4.99\% higher than UCF (0.8267) and 6.35\% higher than FFD (0.8164). Notably, compared to lightweight models like AlexNet, which achieves an average AUC of 0.7903, our method delivers a 7.79\% improvement in average AUC, highlighting its superior performance in terms of accuracy. In terms of efficiency, our model uses only 1.27 GFLOPs and 6.64 million parameters, reducing computation by 87.9\% compared to UCF and 78.9\% compared to FFD, while maintaining superior detection performance.

\begin{figure}
\centering
\includegraphics[scale=0.3]{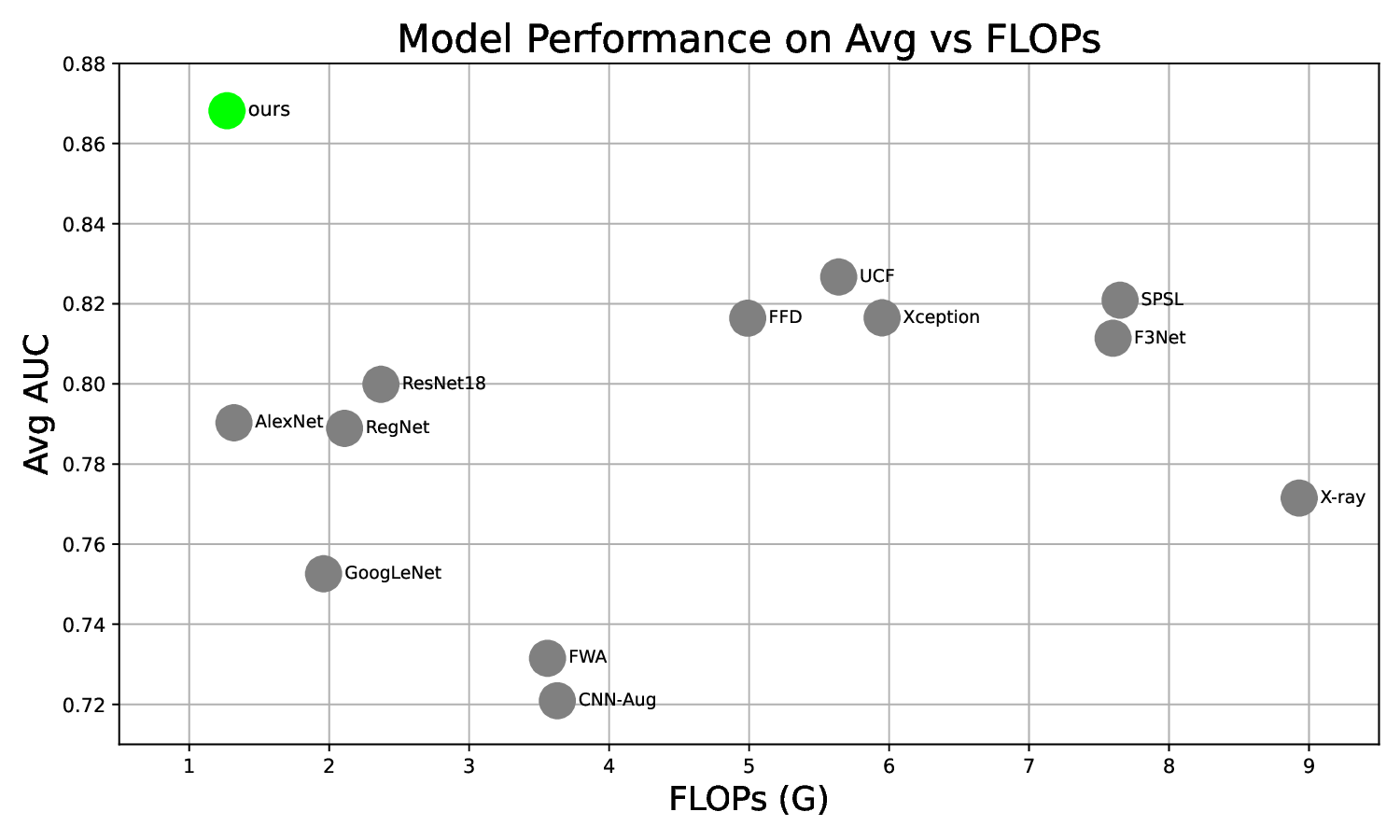}
\caption{Comparison of AUC and FLOPs, highlighting our model's superior performance and efficiency.} \label{fig3}
\end{figure}

To provide a more intuitive comparison, Fig.~\ref{fig3} illustrates the relationship between average AUC and FLOPs. The green dot represents our model (``ours''), which achieves the highest AUC (0.8682) with only 1.27 GFLOPs and 6.64M parameters. In contrast, other high-performing models such as Xception, FFD, and UCF, while offering better AUC, require significantly higher computational resources. Our model strikes a superior balance between performance and efficiency, delivering high detection accuracy while reducing computational cost.

\subsection{Ablation Study}

We conducted ablation experiments to assess the contribution of each module in our model, as shown in Table~\ref{tab:3}. Removing the SFHA module results in significant AUC drops on CDF2 and DFDC, highlighting the importance of spatial-frequency features. Replacing the DBP module with standard max-pooling also degrades performance, demonstrating the superiority of our downsampling approach. Similarly, substituting the TSCA module with simple channel concatenation and convolution reduces accuracy, confirming the effectiveness of token-selective attention. Overall, the complete model consistently outperforms all variants, validating the design choices.

\begin{table}[ht]
\centering
\caption{Ablation study of each module on CDF2 and DFDC datasets.}
\begin{tabular}{cccccc}
\toprule
\textbf{SFHA} & \textbf{DBP} & \textbf{TSCA} & \textbf{CDF2} & \textbf{DFDC} \\
\midrule
\ding{55} & \checkmark & \checkmark & 0.7427 & 0.6779 \\
\checkmark & \ding{55} & \checkmark & 0.7711 & 0.6933 \\
\checkmark & \checkmark & \ding{55} & 0.7832 & 0.7146 \\
\checkmark & \checkmark & \checkmark & \textbf{0.8111} & \textbf{0.7307} \\
\bottomrule
\end{tabular}
\label{tab:3}
\end{table}

\begin{figure}
\centering
\includegraphics[width=\textwidth]{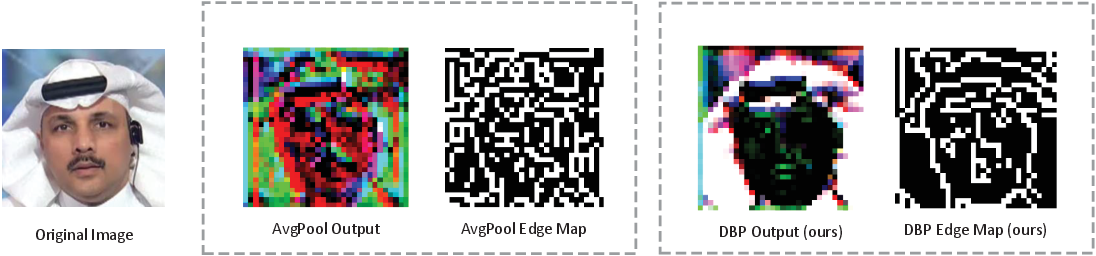}
\caption{Visual comparison of average pooling and our DBP module. DBP preserves clearer edges and structural details.} \label{fig4}
\end{figure}

To demonstrate the effectiveness of our DBP module, Fig.~\ref{fig4} presents an ablation comparison with average pooling. Ablating the DBP module and replacing it with standard average pooling results in blurred features and noisy edges, leading to a loss of important forgery-related details. In contrast, the DBP module preserves sharper structures and clearer boundaries, highlighting its superior capability in retaining crucial forgery cues during downsampling.

\begin{table}[htbp]
\centering
\caption{Ablation study on TSCA: performance vs. computational cost.}
\begin{tabular}{lcccc}
\toprule
\textbf{Model} & \textbf{CDF2} & \textbf{DFDC} & \textbf{FLOPs} & \textbf{Parameters} \\
\midrule
CCA-Fusion & 0.7949 & 0.7216 & 2.28G & 22.19M \\
w/o TSCA & 0.7832 & 0.7146 & \textbf{1.26G} & \textbf{6.46M} \\
\textbf{ours} & \textbf{0.8111} & \textbf{0.7307} & 1.27G & 6.64M \\
\bottomrule
\end{tabular}
\label{tab:4}
\end{table}

To assess the efficiency and effectiveness of the TSCA module, we conducted two ablation studies shown in Table~\ref{tab:4}. The first variant, CCA-Fusion, uses standard cross-attention and increases both FLOPs and parameters significantly. The second, w/o TSCA, removes TSCA and replaces it with basic channel fusion, reducing complexity but lowering accuracy. In contrast, TSCA introduces minimal overhead. FLOPs rise from 1.26G to 1.27G and parameters from 6.46M to 6.64M, while notably boosting performance, with AUC gains on CDF2 from 0.7832 to 0.8111 and on DFDC from 0.7146 to 0.7307. This confirms that TSCA is both lightweight and effective for feature enhancement.

\section{Conclusion}
In this work, we propose SFMFNet, a lightweight and efficient real-time deepfake detection framework. SFMFNet integrates wavelet features and coordinate attention into a dynamic gating mechanism, enhancing the detection of manipulated regions. It also introduces a token-selective cross-attention module to enable efficient cross-scale feature interaction. Additionally, we design a residual downsampling module based on blur pooling, which preserves structural and edge details while reducing information distortion. Extensive experiments demonstrate that SFMFNet achieves an excellent balance between accuracy, speed, and generalization, making it highly suitable for real-world deployment.

%
%
%

\begin{thebibliography}{41}


\bibitem{Simswap}
Chen, R., Chen, X., Ni, B., Ge, Y.: SimSwap: An Efficient Framework for High Fidelity Face Swapping. In: Proceedings of the 28th ACM International Conference on Multimedia (ACM MM), pp. 2003--2011. ACM (2020). \doi{10.1145/3394171.3413630}

\bibitem{StarGANv2}
Choi, Y., Uh, Y., Yoo, J., Ha, J.W.: StarGAN v2: Diverse Image Synthesis for Multiple Domains. arXiv preprint arXiv:1912.01865 (2020). \url{https://arxiv.org/abs/1912.01865}

\bibitem{APSwap}
Wang, T., Li, Z., Liu, R., Wang, Y., Nie, L.: An Efficient Attribute-Preserving Framework for Face Swapping. IEEE Trans. Multimedia 26, 6554--6565 (2024). \doi{10.1109/TMM.2024.3354573}

\bibitem{DeepfakeSurvey2024Wang}
Wang, T., Liao, X., Chow, K.P., Lin, X., Wang, Y.: Deepfake Detection: A Comprehensive Survey from the Reliability Perspective. ACM Comput. Surv. 57(3), Article 58, 35 pages (2024). \doi{10.1145/3699710}


\bibitem{RTDFs}
Mittal, G., Hegde, C., Memon, N.: Gotcha: Real-Time Video Deepfake Detection via Challenge-Response. In: IEEE 9th European Symposium on Security and Privacy (EuroS\&P), pp. 1--20. IEEE (2024). \doi{10.1109/EuroSP60621.2024.00009}


\bibitem{BDD}
Lanzino, R., Fontana, F., Diko, A., Marini, M. R., Cinque, L.: Faster Than Lies: Real-time Deepfake Detection using Binary Neural Networks. In: IEEE/CVF Conf. on Computer Vision and Pattern Recognition Workshops (CVPRW), pp. 3771--3780. IEEE (2024). \doi{10.1109/CVPRW63382.2024.00381}

\bibitem{LaDeDa}
Cavia, B., Horwitz, E., Reiss, T., Hoshen, Y.: Real-Time Deepfake Detection in the Real-World. In: arXiv preprint arXiv:2406.09398 (2024).\url{https://arxiv.org/abs/2406.09398}

\bibitem{inconsistencies}
Li, Y., Chang, M.-C., Lyu, S.: In Ictu Oculi: Exposing AI Created Fake Videos by Detecting Eye Blinking. In: IEEE International Workshop on Information Forensics and Security (WIFS), pp. 1--7. IEEE (2018). \doi{10.1109/WIFS.2018.8630787}


\bibitem{JPEG compression artifacts} Li, Y., Lyu, S.: Exposing Deepfake Videos by Detecting Face Warping Artifacts. In: IEEE Conference on Computer Vision and Pattern Recognition Workshops (CVPRW), pp. 1--5. IEEE (2019). \url{https://arxiv.org/abs/1811.00656}

\bibitem{FF++}
Rössler, A., Cozzolino, D., Verdoliva, L., Riess, C., Thies, J., Nießner, M.: FaceForensics++: Learning to Detect Manipulated Facial Images. In: Proceedings of the IEEE/CVF International Conference on Computer Vision (ICCV), pp. 1--11. IEEE (2019). \doi{10.1109/ICCV.2019.00873}


\bibitem{NoiseDF2023Wang} 
Wang, T., Chow, K. P.: Noise Based Deepfake Detection via Multi-Head Relative-Interaction. In: Proceedings of the AAAI Conference on Artificial Intelligence, vol. 37, no. 12, pp. 14548--14556. AAAI Press (2023). \doi{10.1609/aaai.v37i12.26701}

\bibitem{DCPT2023Wang} 
Wang, T., Cheng, H., Chow, K. P., Nie, L.: Deep Convolutional Pooling Transformer for Deepfake Detection. In: ACM Transactions on Multimedia Computing, Communications, and Applications, vol. 19, no. 6, pp. 1--12. ACM Press (2023). \doi{10.1145/3588574}













\bibitem{extract fine-grained features}
Das, A., Das, S., Dantcheva, A.: Demystifying Attention Mechanisms for Deepfake Detection. In: IEEE International Conference on Automatic Face and Gesture Recognition (FG), pp. 1--7. IEEE (2021). \doi{10.1109/FG52635.2021.9667026}



\bibitem{locate manipulated regions}
Liu, S., Qi, L., Qin, H., Shi, J., Jia, J.: Path Aggregation Network for Instance Segmentation. In: IEEE/CVF Conference on Computer Vision and Pattern Recognition (CVPR), pp. 8759--8768. IEEE (2018). \doi{10.1109/CVPR.2018.00913}



\bibitem{CBAM}
Woo, S., Park, J., Lee, J.Y., Kweon, I.S.: CBAM: Convolutional Block Attention Module. In: Proceedings of the European Conference on Computer Vision (ECCV), pp. 3--19. Springer (2018). \url{https://doi.org/10.1007/978-3-030-01234-2\_1}

\bibitem{wavelet}
De Silva, D. D. N., Vithanage, H. W. M. K., Xavier, S. A., Piyatilake, I. T. S. and Fernando, S.: Parameterized Wavelets for Convolutional Neural Networks. In: IEEE International Conference on Autonomous Robot Systems and Competitions (ICARSC), Ponta Delgada, Portugal, pp. 170--176. IEEE (2020). \doi{10.1109/ICARSC49921.2020.9096125}



\bibitem{BlurPool}
Zhang, R.: Making Convolutional Networks Shift-Invariant Again. In: Proceedings of the International Conference on Machine Learning (ICML), pp. 7324--7334. PMLR (2019). \url{https://arxiv.org/abs/1904.11486}


\bibitem{Celeb-DF_v2}
Li, Y., Yang, X., Sun, P., Qi, H., Lyu, S.: Celeb-DF: A Large-Scale Challenging Dataset for DeepFake Forensics. In: IEEE/CVF Conference on Computer Vision and Pattern Recognition (CVPR), pp. 3204--3213. IEEE (2020). \doi{10.1109/CVPR42600.2020.00327}


\bibitem{DFDC}
Dolhansky, B., Bitton, J., Pflaum, B., Lu, J., Howes, R., Wang, M., Canton Ferrer, C.: The DeepFake Detection Challenge (DFDC) Dataset. arXiv preprint arXiv:2006.07397 (2020). \url{https://arxiv.org/abs/2006.07397}


\bibitem{DFD}
Google AI: Contributing Data to Deepfake Detection Research. \url{https://ai.googleblog.com/2019/09/contributing-data-to-deepfake-detection.html}. Last accessed 9 Jun 2025

\bibitem{DFDCP}
Dolhansky, B., Howes, R., Pflaum, B., Baram, N., Canton Ferrer, C.: The Deepfake Detection Challenge (DFDC) Preview Dataset. arXiv preprint arXiv:1910.08854 (2019). \url{https://arxiv.org/abs/1910.08854}


\bibitem{UADFV}
Li, Y., Chang, M.-C., Lyu, S.: In Ictu Oculi: Exposing AI Generated Fake Face Videos by Detecting Eye Blinking. arXiv preprint arXiv:1806.02877 (2018). \url{https://arxiv.org/abs/1806.02877}


\bibitem{ASAP}
Zhao, Y., Lu, Z., Gong, Y., Song, J., Yang, Y.: ASAP: Advancing Semantic Alignment Promotes Multi-Modal Manipulation Detecting and Grounding. In: Proceedings of the IEEE/CVF Conference on Computer Vision and Pattern Recognition (CVPR), pp. 1--11. IEEE (2025). \url{https://arxiv.org/abs/2412.12718}


\bibitem{AlexNet}
Krizhevsky, A., Sutskever, I., Hinton, G.E.: ImageNet Classification with Deep Convolutional Neural Networks. In: Advances in Neural Information Processing Systems (NeurIPS), vol. 25, pp. 1097--1105. Curran Associates, Red Hook (2012). \doi{10.1145/3065386}

\bibitem{Resnet18}
He, K., Zhang, X., Ren, S., Sun, J.: Deep Residual Learning for Image Recognition. In: Proceedings of the IEEE Conference on Computer Vision and Pattern Recognition (CVPR), pp. 770--778. IEEE, Las Vegas (2016). \doi{10.1109/CVPR.2016.90}


\bibitem{Xception}
Chollet, F.: Xception: Deep Learning with Depthwise Separable Convolutions. In: Proceedings of the IEEE Conference on Computer Vision and Pattern Recognition (CVPR), pp. 1251--1258. IEEE, Honolulu (2017). \doi{10.1109/CVPR.2017.195}

\bibitem{F3Net}
Qian, Y., et al.: Thinking in Frequency: Face Forgery Detection by Mining Frequency‑Aware Clues. In: Proceedings of the European Conference on Computer Vision (ECCV), LNCS, vol. 12357, pp. 86--103. Springer, Glasgow (2020). \doi{10.1007/978-3-030-58610-2\_6}

\bibitem{CNN-Aug}
Wang, S.-Y., et al.: CNN-generated Images are Surprisingly Easy to Spot... for Now. In: Proceedings of the IEEE Conference on Computer Vision and Pattern Recognition (CVPR), pp. 8695--8704. IEEE, Seattle (2020). \doi{10.1109/CVPR42600.2020.00872}

\bibitem{X-rey}
Li, L., et al.: Face X-ray for More General Face Forgery Detection. In: Proceedings of the IEEE Conference on Computer Vision and Pattern Recognition (CVPR), pp. 5000--5009. IEEE, Seattle (2020). \doi{10.1109/CVPR42600.2020.00505}

\bibitem{FFD}
Dang, H., et al.: On the Detection of Digital Face Manipulation. In: Proceedings of the IEEE Conference on Computer Vision and Pattern Recognition (CVPR), pp. 5781--5790. IEEE, Seattle (2020). \url{https://arxiv.org/abs/1910.01717}

\bibitem{UCF}
Yan, Z., Zhang, Y., Fan, Y., Wu, B.: UCF: Uncovering Common Features for Generalizable Deepfake Detection. In: Proceedings of the IEEE Conference on Computer Vision and Pattern Recognition (CVPR), pp. 12454--12463. IEEE, Virtual (2023). \doi{10.1109/CVPR52729.2023.01210}

\bibitem{RegNet}
Xu, J., Pan, Y., Pan, X., Hoi, S., Zhang, Y., Xu, Z.: RegNet: Self-Regulated Network for Image Classification. IEEE Transactions on Neural Networks and Learning Systems, 34(11), 9562--9567 (2023). \doi{10.1109/TNNLS.2022.3158966}

\bibitem{GoogLeNet}
Szegedy, C., Liu, W., Jia, Y., Sermanet, P., Reed, S., Anguelov, D., Erhan, D., Vanhoucke, V., Rabinovich, A.: Going Deeper with Convolutions. In: Proceedings of the IEEE Conference on Computer Vision and Pattern Recognition (CVPR), Boston, MA, USA, pp. 1--9 (2015). \doi{10.1109/CVPR.2015.7298594}



\bibitem{FWA}
Li, Y., Lyu, S.: Exposing DeepFake Videos by Detecting Face Warping Artifacts. arXiv preprint arXiv:1811.00656 (2018). \url{https://arxiv.org/abs/1811.00656}

\bibitem{Capsule}
Nguyen, H.H., Yamagishi, J., Echizen, I.: Capsule-Forensics: Using Capsule Networks to Detect Forged Images and Videos. In: ICASSP 2019 - 2019 IEEE International Conference on Acoustics, Speech and Signal Processing (ICASSP), Brighton, UK, pp. 2307--2311. IEEE (2019). \doi{10.1109/ICASSP.2019.8682602}

\bibitem{SPSL}
Liu, H., Li, X., Zhou, W., Chen, Y., He, Y., Xue, H., Zhang, W., Yu, N.: Spatial-Phase Shallow Learning: Rethinking Face Forgery Detection in Frequency Domain. In: Proceedings of the IEEE/CVF Conference on Computer Vision and Pattern Recognition (CVPR), Nashville, TN, USA, pp. 772--781 (2021). \doi{10.1109/CVPR46437.2021.00083}

\bibitem{DeepfakeBench}
Yan, Z., Zhang, Y., Yuan, X., Lyu, S., Wu, B.: DeepfakeBench: A Comprehensive Benchmark of Deepfake Detection. In: Advances in Neural Information Processing Systems (NeurIPS) Datasets and Benchmarks Track, vol. 36 (2023). \url{https://arxiv.org/abs/2307.01426}








\end{thebibliography}
%

\end{document}